\documentclass[journal]{IEEEtran}
\usepackage{blindtext}
\usepackage{graphicx}
\usepackage{subfig}
\usepackage{multirow}
\newcommand{\rx}{$\times$}
\usepackage{verbatim}
\usepackage{amsmath}

\ifCLASSINFOpdf
\else
\fi
\begin{document}

\title{On Low-Resolution Face Recognition in the Wild: Comparisons and New Techniques}

\author{Pei Li${}^1$\hspace{0.5in}  Loreto Prieto${}^2$\hspace{0.5in} Domingo Mery${}^2$\hspace{0.5in} Patrick J. Flynn${}^1$\\

${}^1$Department of Computer Science and Engineering\\University of Notre Dame\\
${}^2$Department of Computer Science\\Pontificia Universidad Cat\'{o}lica de Chile}

\maketitle

\begin{abstract}
Although face recognition systems have achieved impressive performance in recent years, the low-resolution face recognition task remains challenging, especially when the low-resolution faces are captured under non-ideal conditions, as is common in surveillance-based applications. Faces captured in such conditions are often contaminated by blur, non-uniform lighting, and non-frontal face pose. In this paper, we analyze face recognition techniques using data captured under low-quality conditions in the wild. We provide a comprehensive analysis of experimental results for two of the most important applications in real surveillance applications, and demonstrate practical approaches to handle both cases that show promising performance. The following three contributions are made:
{\em (i)} we conduct experiments to evaluate super-resolution methods for low-resolution face recognition;
{\em (ii)} we study face re-identification on various public face datasets including real surveillance and low-resolution subsets of large-scale datasets, present a baseline result for several deep learning based approaches, and improve them by introducing a Generative Adversarial Network (GAN) pre-training approach and fully convolutional architecture; and 
{\em (iii)} we explore low-resolution face identification by employing a state-of-the-art supervised discriminative learning approach. Evaluations are conducted on challenging portions of the SCface and UCCSface datasets.

\end{abstract}


\section{Introduction}
\label{Sec:Introduction}
In recent years, we have witnessed tremendous improvements in face recognition system performance, especially when employing different specially designed deep learning architectures. Some state-of-the-art methods based on deep learning models including \cite{sun2014deep}, \cite{schroff2015facenet}, \cite{parkhi2015deep}, \cite{sun2016sparsifying}, \cite{wen2016discriminative} and \cite{liu2017sphereface} have achieved an accuracy over 99\% on public face datasets such as LFW \cite{LFWTech}. Although these algorithms can deal effectively with faces with significant pose variations, these faces generally need to be large in area. Also, pre-processing techniques such as face frontalization and face alignment are needed. These procedures, that are designed for HR face image data, cannot be applied directly to low-quality face images. The use of surveillance systems in a wide range of public places is increasing, creating a challenging use case for face recognition in an environment where detected faces will be in low-resolution (LR). We refer to this as the {\em Low-Resolution Face Recognition} (LRFR) problem. Practical face recognition systems for images captured in surveillance scenarios can address face identification tasks (using a watch list) and face re-identification tasks (where a subject is matched to a previous appearance in a surveillance system).

This work is based in part on prior work published in \cite{li2017learning}, \cite{li2016toward} but presents many new contributions. In the original work \cite{li2017learning}, we focus on prototyping lightweight deep neural networks for LR face re-identification and showed that LR face images can be used for person re-identification. We employed data from a real surveillance system and conducted an extensive person re-identification study. We address the LR face recognition task in a larger scope which includes a qualitative exploration of the recognition performance gap between controlled facial images and video quality face images using super-resolution techniques, face identification, and face re-identification. We also focus on boosting the performance based on \cite{li2017learning} and extend our evaluations using more datasets including SCface, UCCSface and the MegaFace challenge 2 LR subset.

In Section \ref{Sec:RelatedWorks}, we provide a literature review and summary descriptions of several newly released real surveillance datasets. In Section \ref{Sec:Methods_and_Experiments}, we present new methods and experiments as follows: 
Section \ref{sec:sr_evaluation} presents several baseline face recognition results obtained by evaluating state-of-the-art super-resolution methods \cite{liu2016robust,yang2010image}.
A summary of the performance of super-resolution augmented face recognition techniques employing LR face inputs from two popular face datasets (AR \cite{martinez1998ar} and YouTube Faces (YTF) \cite{wolf2011face}) is also presented, and illustrates the performance gap between recognition of faces captured in a controlled environment and recognition of faces captured in the wild.
In Section \ref{sec:lr_faceidentification}, we propose a center regularization approach for LR face recognition. The goal of this work is to learn a common feature space that locates LR and HR face images of the same subject as close to one another as possible in feature space, without generating thousands of pairs for training. By conducting comprehensive evaluations on two different datasets, we are able to show the performance gap between open-set and closed-set face identification when employing LR inputs.
In Section \ref{sec:lr_face_reid}, we first summarize the discoveries in our previous work \cite{li2017learning} based on a real surveillance dataset named VBOLO. Different deep architectures are designed and evaluated for LR surveillance face images as a baseline result and are improved by introducing fully convolutional spatial pyramid pooling (SPP). To generalize the ability of the performance on larger data scale, we further study face re-identification on several surveillance and LR datasets from videos or images collected online. In addition, we present a novel DCGAN-based pre-training approach to further boost performance.
In Section \ref{Sec:Conclusions}, concluding remarks are provided. 
\section{Related Work and Datasets}
\label{Sec:RelatedWorks}
In this section, we present related works on face recognition for LR images and face re-identification, and describe some important datasets that can be used in this area of research.
\subsection{Low-Resolution Face Recognition}
The LRFR task is a subset of the general face recognition problem.
One of the most useful application scenarios for this task is recognition using surveillance imagery. In this scenario, faces are captured from cameras with a large standoff, positioned above head height, and sometimes under challenging lighting conditions. Even if the faces could be detected by a specially trained face detector, it is hard to construct a robust feature representation due to the lack of information embedded in the image itself. Although HR face recognition systems have achieved nearly perfect performance on several datasets, LRFR remains a challenging problem. Approaches to this problem in the literature follow two main themes. Some techniques employ super-resolution (SR) and deblurring techniques to increase the input LR face size to a point where a high-resolution (HR) face recognition technique may work well. Another popular approach is to learn a unified feature space for LR and HR face images, within which feature vector distance plays its typical role as a matching score.

SR techniques are widely used to turn LR images into better quality images. Face SR approaches are thus an intuitive way to recover LR face images for recognition. Hennings-Yeomans et al. \cite{2008:Hennings} included the prior extracted face features as measures of fit of the SR result, and performs SR from both reconstruction and recognition perspectives. Yang et al. \cite{yang2010image} proposed a joint dictionary training method for general SR that employs both LR and HR image patches. Zou et al. \cite{2013:Zou} designed a linear regression model with two elements (new data and discriminative constraints) to learn the mapping. Jiang et al. \cite{2014:Jiang} proposed a coarse-to-fine face SR approach via a multi-layer locality-constrained iterative neighbor
embedding. Kolouri et al. \cite{Kolouri_2015_CVPR} introduced a single frame SR technique that uses a transport-based formulation of this problem. Wang et al. \cite{wang2016studying} deployed deep learning pre-training with a carefully selected loss function to achieve SR for matching between LR and HR face images, and achieves state-of-the-art performance on a new surveillance dataset. 

Another approach to solving the LR face recognition problem is to seek a unified feature space that preserves proximity between faces of different resolutions. Li et al. \cite{2010:Li} proposed a coupled mapping method that projects face images with different resolutions into a unified feature space. Biswas et al. \cite{biswas2010multidimensional} simultaneously embedded the LR and HR faces in a common space such that the distances between them in the transformed space approximates the distance between between two HR face images of the same subject. Ren et al. \cite{ren2012coupled} used a coupled mapping strategy with both HR and LR counterparts for learning the projection directions as well as exploiting discriminant information. Shekhar et al. \cite{2011:Shekhar} proposed robust dictionary learning for LR face recognition that shares common sparse codes.  The technique of Qiu et al. \cite{qiu2014dictionary} learned a domain adaptive dictionary to handle the matching of two faces captured in source and target domains. Li et al. \cite{li2016toward} proposed several shallow network structures to learn a latent space between LR and HR images, and evaluated the proposed methods on a new surveillance dataset. There are also some other methods such as  \cite{rahtu2012local} and  \cite{herrmann2016low}, which explore more robust features directly to improve recognition rate in blurry and degraded face images. In addition, methods like those proposed in  \cite{2011:Haichao},  \cite{2011:Nishiyama} and  \cite{2014:Pan} work to restore the upsampled LR images using deblurring techniques.

In the last three years, novel face recognition methods based on deep learning for low-quality face images have been developed. The most relevant approaches are as follows.
In \cite{wang_2016}, partially coupled networks are proposed for unsupervised super-resolution pre-training. The classifier is obtained by fine-tuning on a different dataset for specific domain simultaneous super-resolution and recognition.
In \cite{juefeixu_2016}, \cite{juefeixu_2017}, an attention model that shifts the network's attention during training by blurring the images with various percentage of blurriness is presented for gender recognition. In \cite{mcpherson_2016}, three obfuscation techniques are proposed to restore face images that have been degraded by {\em mosaicing} (pixelation) and blurring processes.
In \cite{bulat_2018}, a multi-task deep model is proposed to simultaneously learn face super-resolution and facial landmark localization. The face super-resolution subnet is trained using a generative adversarial network (GAN) \cite{creswell_2018}, \cite{ledig_2017} (see a comparison of different versions of GANs in the context of face super-resolution in \cite{chen_2017_gan}). In \cite{huang_2017}, inspired by the traditional wavelet that can depict the contextual and textural information of an image at different levels, a deep architecture is proposed. In \cite{chen_2018}, a network that contains a coarse super-resolution network to recover a coarse HR image is presented. It is the first deep face super-resolution network utilizing facial geometry prior to end-to-end training and testing.
In \cite{chrysos_2017}, a network for deblurring facial images using a Resnet-based non-max-pooling architecture is proposed. In \cite{yu_2018}, a face hallucination method based on an upsampling network and a discriminative network is proposed. The approach includes feature maps with additional facial attribute information. In \cite{shen_2018}, global semantic priors of the faces are exploited in order to restore blurred face images.  In \cite{lu2018deep}, a new branch network that can be appended to a trunk network to match a different resolution of probe images to the gallery images was designed.   

In all these methods, we see that automatic face recognition is far from perfect when tackling more challenging images of faces taken in unconstrained environments, {\em e.g.} surveillance, forensics, etc. Most of the approaches mentioned above are evaluated on low-quality versions of constrained face datasets like Multi-PIE, FERET or FRGC. The images are created by directly downsampling or blurring original images. However, the LR face recognition problem becomes a challenge when faces captured in an unconstrained environment. In such cases, the LR face recognition problem needs to be explored more extensively.

\subsection{Face Re-Identification}
\label{sec:face_reid}
Here, we briefly introduce person re-identification (ReID) and our motivation for face re-identification.
A typical end-to-end ReID system generally involves the following steps: person detection, preprocessing, feature extraction, and matching. It is widely used for surveillance purposes, modeled as given a pair of images. The objective is to find whether the images are from the same person or not. Often, the pictures are captured from different cameras in a surveillance network.
Traditional approaches to the ReID problem focus on two main components: feature extraction and similarity computation for matching. We can divide most of the existing methods into two categories: methods employing deep learning, and those without. For the non-deep learning methods, research have proposed and used different handcrafted features such as symmetry-driven accumulation of local features (SDALF) \cite{Bazzani:CVIU13}, color histograms \cite{xiong2014person,zhang2014novel}, color names \cite{yang2014salient}, local binary patterns \cite{khamis2014joint}, aggregation of patch summary features \cite{zhao2013person}, metric learning approaches \cite{xiong2014person,khamis2014joint,koestinger2012large,li2013locally,martinel2014saliency,li2013learning,xiong2014person,jose2016scalable,chen2016similarity,zhang2016learning} and various combinations of these.  Several deep learning approaches use ``Siamese''  deep convolutional neural networks \cite{yi2014deep,varior2016gated,varior2016siamese,liu2016end} for feature extraction and metric learning at the same time providing novel end-to-end solutions. Ahmed et al. \cite{ahmed2015improved} proposed a new deep learning framework based on the idea from Yi et al. \cite{yi2014deep},  where two novel layers are employed for computing cross-input neighborhood differences by integrating local relationships based on mid-level features. They additionally showed that the features acquired from the head and neck could be an important clue for person re-identification. Wu et al. \cite{wu2016personnet} improved performance based on Ahmed's idea by using a deeper architecture and a new optimization method. Other deep network structures such as \cite{ustinova2015multiregion} and \cite{subramaniam2016deep} have been designed which also effectively solved the ReID problem on older ReID datasets.
Qui et al. \cite{qiu2014dictionary} attempted to perform facial ReID by using domain adaptation methods to reconcile different facial poses; however, their experiments were performed on the Multi-PIE \cite{gross2010multi} dataset, in which face images have controlled poses and illuminations.
 Although an increasing number of surveillance cameras have been deployed in public areas, the quality of the video frames is usually low and people captured in the frame are in an uncontrolled pose and illumination condition. Thus, general person re-identification can be a challenging task. As \cite{burton1999face} shows, body and gait might play a role in recognizing the target in LR video frames, however, obscuring the target produced a dramatic drop in human-level recognition performance. Also, the face-mask-out experiment in \cite{li2016toward} also demonstrates that the face could be an indispensable part of identity recognition. These works help to motivate the LRFR problem as a component of the re-identification problem.

\subsection{Datasets}
There are some good surveillance datasets and also some large scale unconstrained face datasets that contain natural LR faces suitable for training and testing LRFR systems.

\label{sec:dataset}
Most of the LR face images used for research are generated by downsampling a standard face recognition dataset that is collected in a controlled environment.  We select the AR dataset to research the LRFR task under uncontrolled scenarios and other unconstrained LR face datasets for more exploration.    
It is used to illustrate the different data distribution between LR face images artificially generated from high quality controlled face images and those directly collected in unconstrained scenarios such as surveillance camera networks.

\subsubsection{AR Face}
The images in the AR database \cite{martinez1998ar} were taken from 100 subjects (50 women and 50 men) with different facial expressions, illumination conditions, and occlusions by sunglasses or scarves under strictly controlled conditions. In our work, these images are used to estimate the baseline performance of face recognition methods on aligned face images in controlled environments.

\subsubsection{MegaFace Challenge 2 LR subset}
The MegaFace challenge 2 (MF2) training dataset \cite{2017:Nech} is the largest (in the number of identities) publicly available facial recognition dataset, with 4.7 million face images and over 672,000 identities. The MF2 dataset is obtained by running the Dlib \cite{king2009dlib} face detector on images from Flickr  \cite{thomee2016yfcc100m}, yielding 40 million unlabeled faces across 130,154 distinct Flickr accounts. Automatic identity labeling is performed using a clustering algorithm.
We performed a subset selection from the MegaFace Challenge 2 training set with tight bounding boxes to generate a LR subset of this dataset. Faces smaller than 50x50 pixels are gathered for each identity, and then we eliminated identities with fewer than five images available. This subset selection approach produced 6,700 identities and 85,344 face images in total. The extraction process does yield some non-face images, as does the original dataset processing. No further data cleaning is conducted on this subset.


\subsubsection{YouTubeFaces}
The YouTubeFaces Database  \cite{wolf2011face} is designed for studying the problem of unconstrained face recognition in video. The dataset contains 3,425 videos of 1,595 different people. An average of 2.15 videos is available for each subject.
\subsubsection{SCface}
SCface \cite{grgic2011scface} is a database of static images of human faces collected in an uncontrolled indoor environment using five video surveillance cameras of various qualities. The database contains 4,160 static images (in the visible and infrared spectra) of 130 subjects. We choose the HR and LR visible face subset for training and testing.

\subsubsection{UCCSface}
The UnConstrained College Students (UCCS) dataset \cite{sapkota2013large} contains HR images captured from an 18-megapixel camera at the University of Colorado at Colorado Springs, capturing people walking on a campus sidewalk from a standoff of 100 to 150 meters, at one frame per second. The dataset consists of more than 70,000 hand-cropped face regions. An identity is manually assigned to many of the faces. We use the database subset that has assigned identities (180 identities total). Although the data are captured by high-definition cameras, the face regions are tiny due to the large standoff and contain a lot of noise and blurriness.

\subsubsection{VBOLO face}
This dataset \cite{li2016toward}, \cite{li2017learning} (formerly known as EBOLO) was collected in several sessions at various checkpoints within public transportation facilities such as tunnels, bridges, and hallways. These capture environments include different camera mount heights and depression angles, illuminations, backgrounds, resolutions, pedestrian poses, and distractors. This dataset provides a good scenario for the facial ReID problem. This dataset uses a small set of known individuals (``actors'') who move in and out of the surveillance cameras' fields of view, together with unknown persons (``distractors'').  The actors change clothing randomly between each appearance in a camera's field of view.
Compared to a typical body-based ReID dataset, which has only a few images for each subject, the VBOLO dataset has a large number of annotations for each subject from consecutive video frames, which mimic a real scenario for surveillance tracking and detection. This is significantly challenging for matching, because
{\em i)} faces change size significantly and exhibit significant pose variations; and
{\em ii)} the two cameras supplying the probe and gallery images may have different resolutions and points of view. In VBOLO, we employ videos captured at two distinct locations (denoted as Station 1 and Station 2) in this research. Each of the collections has nine actors with nine appearances each. 

\section{Methods and Experiments}
\label{Sec:Methods_and_Experiments}
In this Section, we present four groups of methods and experiments on LRFR: {\em A)} super-resolution techniques, {\em B)} comparison between virtual and real LR, {\em C)} face identification and {\em D)} face re-identification.

\subsection{Super-resolution Techniques}
\label{Sec:SR_Techniques}
\subsubsection{Description}
In order to explore the gap between the constrained and unconstrained LR face recognition performance, we designed a small super-resolution (SR) experiment with the AR \cite{martinez1998ar} and YouTube Faces (YTF) \cite{wolf2011face} datasets.
In this experiment, the idea is to evaluate the matching performance of two face images: a LR image and a HR image.

\begin{figure}[t!]
\begin{center}
\includegraphics[width=\columnwidth]{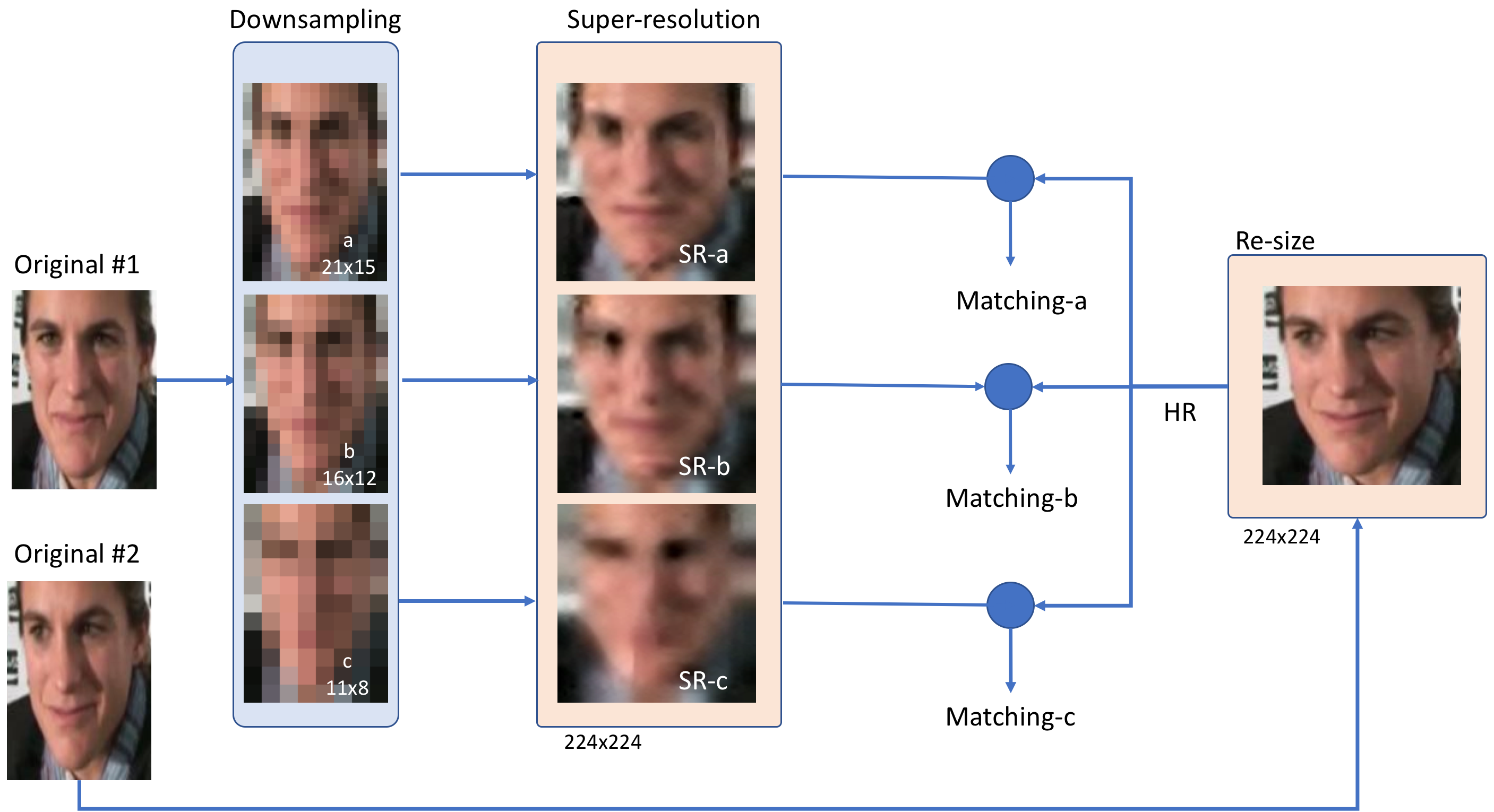}
\caption{Super-resolution experiment: two images, \#01 (for LR purposes) and  \#02  (for HR purposes), are matched using five different super-resolution algorithms for two different LR sizes.}
\label{Fig:lr_sr_experiment}
\end{center}
\end{figure}

In the data selection process, two images or video frames are selected from the datasets for each subject. As illustrated in Fig. \ref{Fig:lr_sr_experiment}, `Original \#1' and  `Original \#2' are used for LR and HR purposes respectively. The first one is downsampled to each of three LR sizes:  (a) 21$\times$15, (b) 16$\times$12, and (c) 11$\times$8 pixels. The second one is upsampled to the matcher's required input size of 224 $\times$ 224 pixels. In both cases, bicubic interpolation is used \cite{2009:Gonzalez}. The images or frames are selected as follows.

\noindent $\bullet$ {\bf AR dataset}: original LR and HR face images are taken from face images \#01 and \#14 respectively of each subject. Since the AR dataset has 100 subjects, we obtained 100 LR--HR pairs of face images for this experiment.

\noindent $\bullet$  {\bf YouTube Faces dataset}: original LR and HR face images are chosen as the earliest and latest frame containing a frontal face image in the latest video clip available for each subject. The detection of frontal faces is performed by applying Zhu's approach \cite{2012:Zhu} to each frame and selecting face images where Zhu's method returned a pose of 0. The dataset has 1,595 subjects; however, only 1,463 subjects have two suitable frames available in the last video clip. Thus, we have 1,463 LR--HR pairs of face images available.

After the selection of images and frames and scaling to target HR and LR sizes is performed, each of the LR images from that process is then upscaled to the same size of the HR image (224$\times$224 pixels) by each of the following methods: (a) bicubic interpolation \cite{2009:Gonzalez}, (b) SCN \cite{liu2016robust}, (c)
sparse representation super-resolution (ScSR) \cite{yang2010image}, (d) LapSRN \cite{LapSRN}, and (e) SRGAN \cite{ledig2017photo}. In addition, we included in the experiments the `Direct method', where original images \#1 and \#2 are compared without downsampling using bicubic interpolation only to achieve the size of 224$\times$224 pixels required by the matcher. This experimental design is depicted in Fig. \ref{Fig:lr_sr_experiment}.

\subsubsection{Experiments and Results}
The experimental approach is presented in Fig. \ref{Fig:lr_sr_experiment}. The VGG-face trained network in \cite{parkhi2015deep} is used to produce a feature vector for the HR image and all fifteen of the SR images.  Matching scores are obtained for each match and nonmatch pair involving one HR and one of the fifteen different upscaled SR images. The cosine distance is used as a match score. 

\begin{figure}[t!]
\begin{center}
{\includegraphics[width=3.2 in]{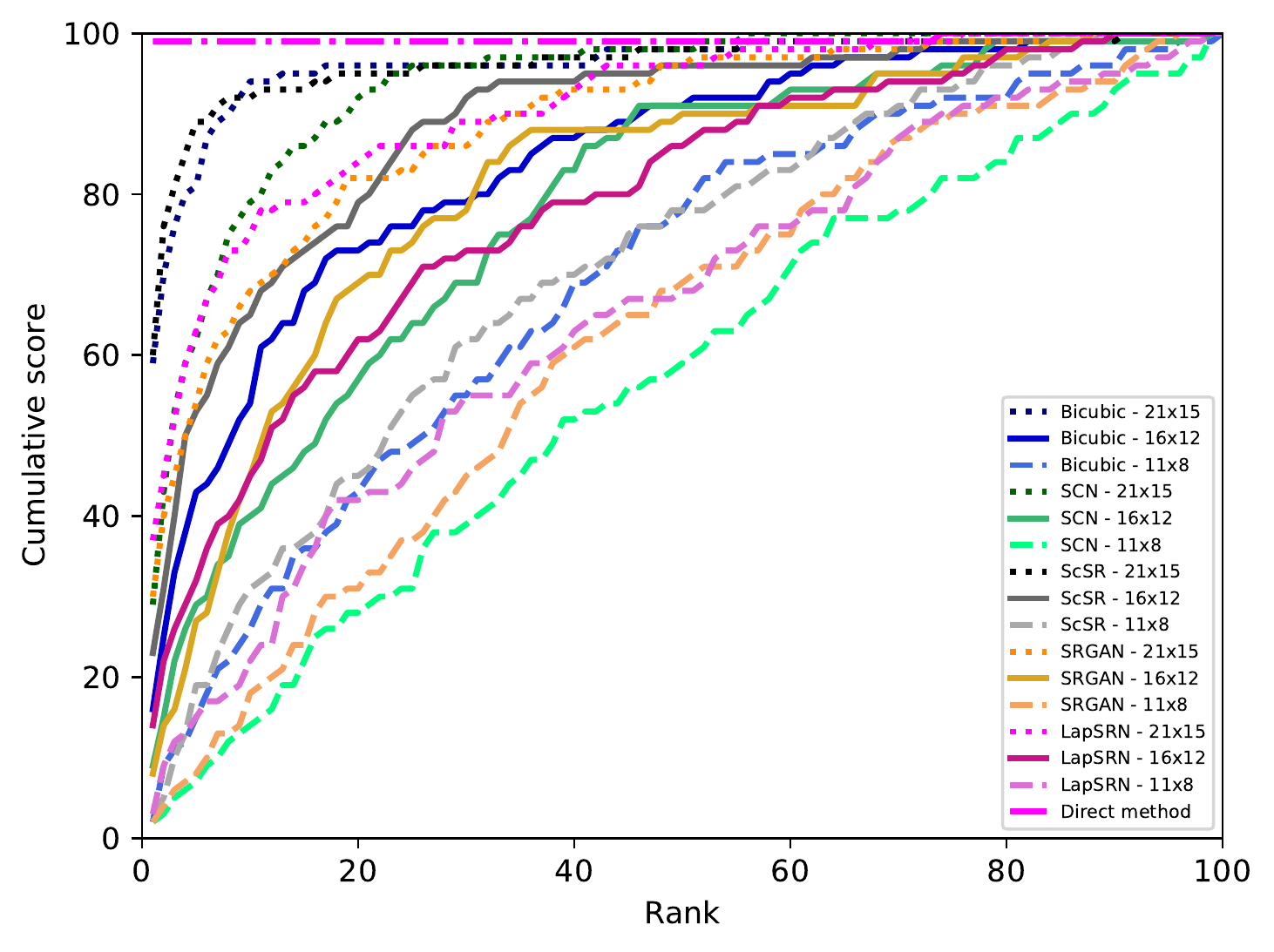}}
\caption{Cumulative match characteristic for AR.}
\label{Fig:SR-AR}
\end{center}
\end{figure}

\begin{figure}[t!]
\begin{center}
{\includegraphics[width=3.2 in]{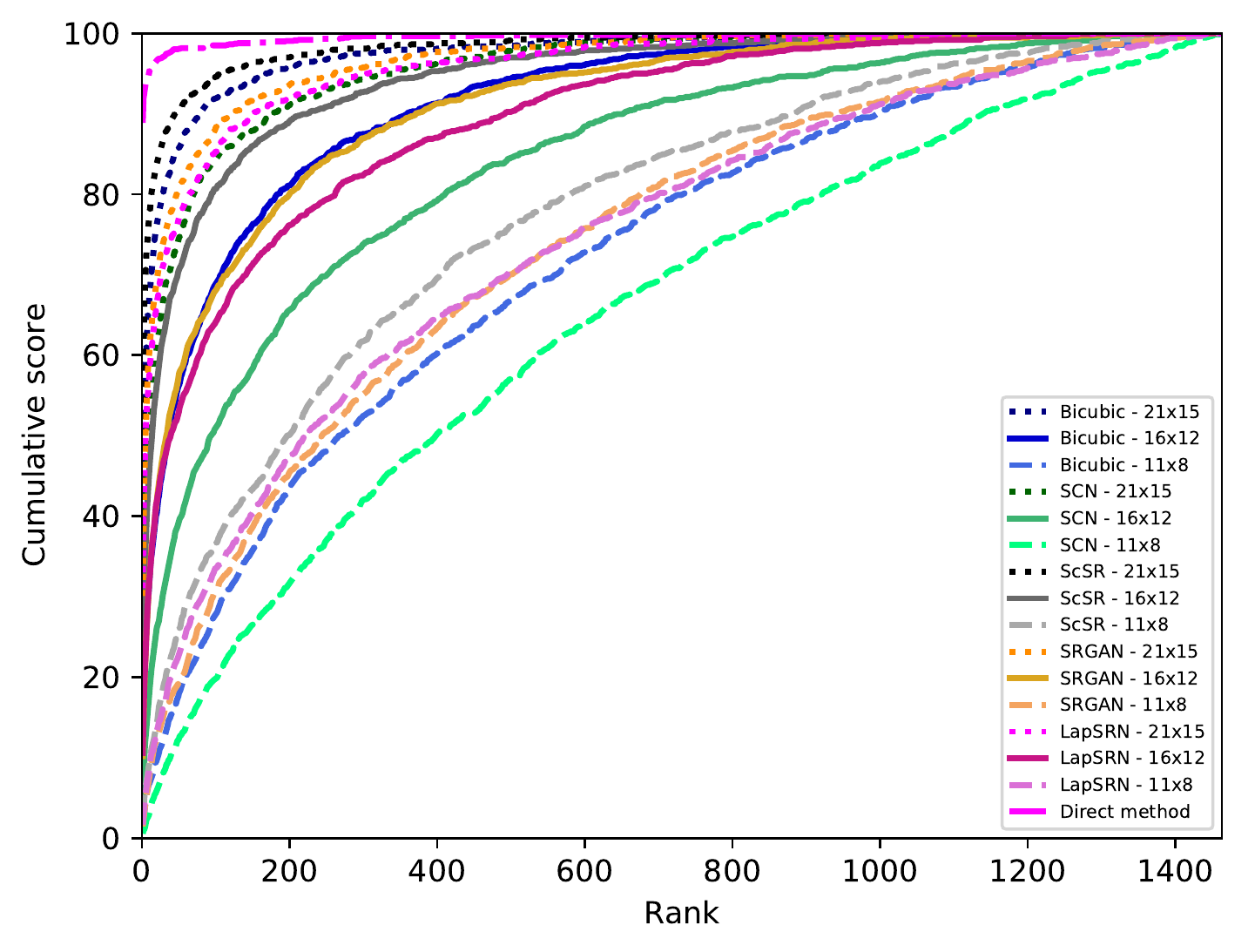}}
\caption{Cumulative match characteristic for YouTube Faces.}
\label{Fig:SR-YT}
\end{center}
\end{figure}

The cumulative match characteristic curve for the AR and YouTube Faces datasets can be seen in Figs. \ref{Fig:SR-AR} and \ref{Fig:SR-YT} respectively. As might be expected, the performance decreases with decreasing resolution: the Direct method (with no downsampling) achieves better results than those obtained by LR images. Moreover, 21$\times$15 LR-images obtain better performance than 16$\times$12 LR-images, and these ones better than 11$\times$8 LR-images. 

In order to show the degradation of performance with low-quality images, we conducted another experiment. Since the original faces in YouTube video frames might be of low quality, we evaluate the performance not only in the matching of all 1,463 pairs, but also in the 500 pairs with the highest quality and in the 500 pairs with lowest quality. We call these two new subsets HQ-YT and LQ-YT respectively. For the measurement of quality of the original face images, we use a score based on the ratio between the high-frequency coefficients and the low-frequency coefficients of the wavelet transform of the image \cite{2013:Pertuz}. Low score values indicate low quality. We evaluate the performance at rank 10 when changing the downsampling target resolution in HQ-YT and LQ-YT. This can be seen in Table \ref{Tab:HQ-LQ}, where high-quality images yield a significantly better performance than the low-quality ones. 

The results of Figs. \ref{Fig:SR-AR} and \ref{Fig:SR-YT} demonstrate that, for these datasets and these algorithm implementations and at the same resolution, sparse representation super-resolution (ScSR) and bicubic interpolation consistently outperform deep learning super-resolution methods. The poor performance of the deep learning method is due to the introduction of artifacts in severely degraded images as we can see in the example of Fig. \ref{Fig:lr_sr_experiment}. 

\begin{table}[t!]
\centering
\caption{Rank-10 correct match score for subsets HQ-YT and LQ-YT}
\normalsize
\begin{tabular}{llccc}
\hline
\hline
\multicolumn{2}{l}{Subset } & (a)  & (b)  & (c)   \\ 
\multicolumn{2}{l}{} & 21$\times$15  & 16$\times$12  & 11$\times$8   \\ 
\hline
HQ-YT:                  & Bicubic 
            & 80.8\% & 53.2\% & 15.6\% \\  
                        & SCN 
            & 69.6\% & 35.2\% &  9.6\% \\  
                        & ScSR 
            & {\bf 87.2\%} & {\bf 66.4\%} & {\bf 26.2\%} \\ 
                        & LapSRN
            & 78.4\% & 56.4\% &  20.4\% \\  
                        & SRGAN 
            & 75.4\% & 53.4\% &  16.2\% \\  
\hline
LQ-YT:                  & Bicubic 
            & 81.4\% & 45.8\% & 10.8\% \\ 
                        & SCN 
            & 64.4\% & 27.2\% &  7.2\% \\  
                        & ScSR 
            & {\bf 87.2\%} & {\bf 61.0\%} & {\bf 14.6\%} \\ 
                        & LapSRN
            & 64.7\% & 38.9\% &  9.8\% \\  
                        & SRGAN 
            & 70.9\% & 40.5\% &  12.4\% \\ 
\hline
\hline
\end{tabular}
\label{Tab:HQ-LQ}
\end{table}

\subsection{Comparison between FR on virtual and real LR images }
\label{sec:sr_evaluation}
\subsubsection{Description}
In order to explore the gap between virtual (synthetic) LR and real LR, we designed an experiment with the YouTube Faces and SCface datasets to evaluate the gap in performance by matching a HR image and a LR image scaled up with either bicubic or SRGAN between a synthetic LR dataset (computed from YouTube Faces as explained in Section \ref{Sec:SR_Techniques}) and a real LR dataset (SCface). The upscaling methods used are a traditional method (bicubic) and a deep learning method (SRGAN).

\noindent $\bullet$  {\bf YouTube Faces dataset}: The same image selection as the super-resolution experiment is used. Then the LR images are downsampled to 16$\times$12 using bicubic interpolation.

\noindent $\bullet$  {\bf SCface}: For each of the 130 subjects in the dataset, the HR mugshot and the LR taken by camera 3 at a standoff of 4.20 meters are chosen as gallery images and probe images. The images are resized to 16$\times$12 as well. Then SRGAN and bicubic interpolation are employed to output face images of 224$\times$224 pixels.

\subsubsection{Experiments and Results}
\begin{figure}
\begin{center}
{\includegraphics[width=3.2 in]{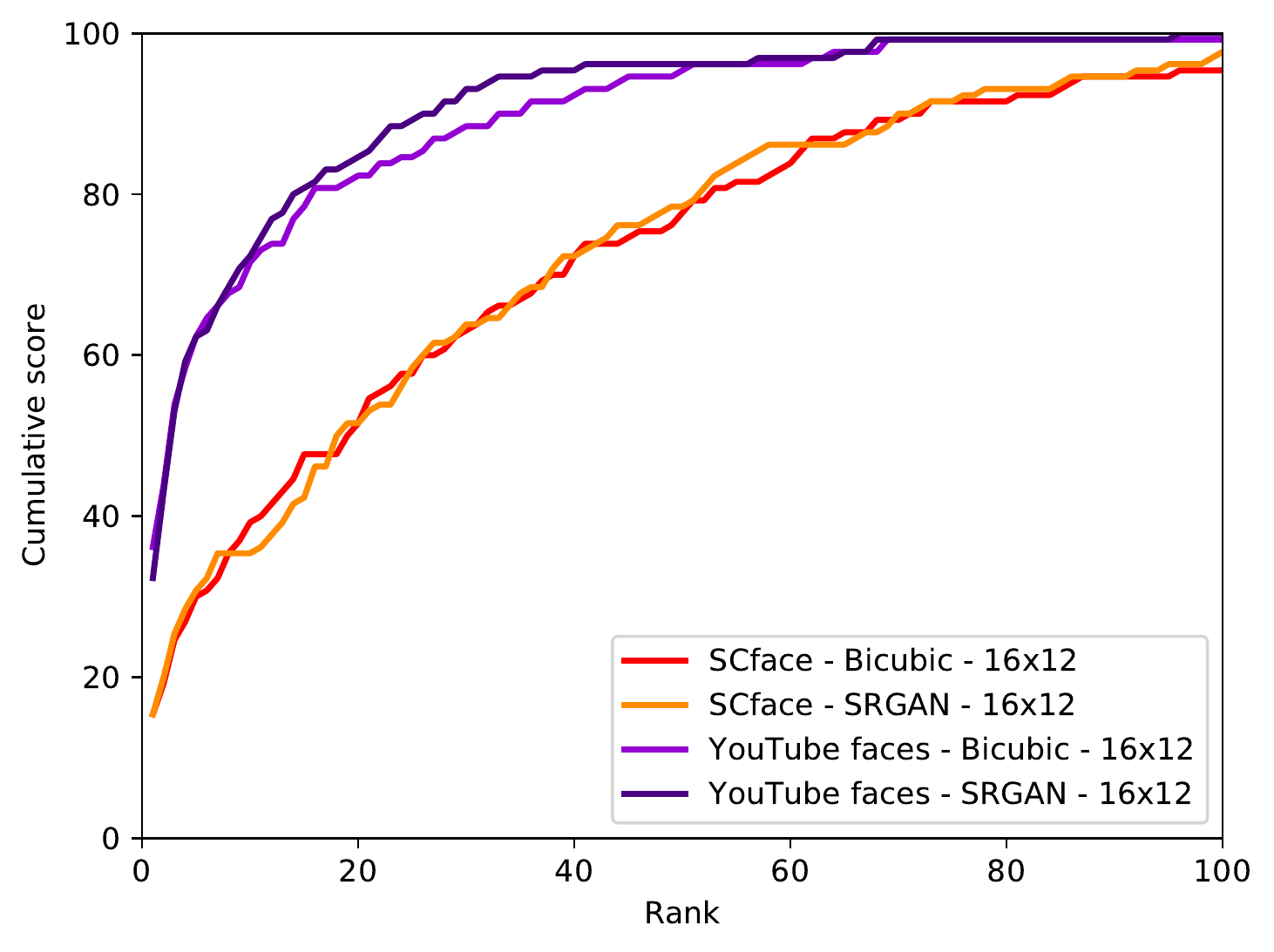}}
\caption{Cumulative match characteristic plot for virtual and real LR experiment}
\label{Fig:Vir-Real}
\end{center}
\end{figure}

The experimental approach is similar to the super-resolution experiment. We use a trained VGG-face network \cite{parkhi2015deep} to produce feature vectors for the HR and LR images for both datasets and scaling methods. The cosine distance is used as a score. The cumulative match characteristic curve for 100 random subjects from Youtube Faces and SCface and both methods can be seen in \ref{Fig:Vir-Real}. 

The decision to use 100 random subjects is taken to make the comparison fairer. If all the subject are used in order to get a 100\% performance in rank-1 YouTube Faces would need to get 1463 subjects correctly but SCface only 130 subjects.

Fig. \ref{Fig:Vir-Real} shows that the performance of the super-resolution approaches on virtual LR is consistently much better than the real LR. Hence, in other to characterize system performance on real LR images, real LR face images should be used instead of synthetic LR face images obtained by simply downsampling the HR face images.

\subsection{Low-Resolution Face Identification}
\label{sec:lr_faceidentification}


In this section, we focus on LR face identification. We first focus on cross-resolution face identification which applies when the enrolled face images are mostly collected in controlled scenarios with HR and LR faces are captured with surveillance cameras with an uncontrolled pose and lighting conditions. This is a challenging recognition task which relies strongly on a good resolution invariant representation.



\subsubsection{Description}

Most existing approaches to cross resolution matching would learn a unified space in which LR and HR faces are represented. This requires a carefully designed face pairing mining strategy during the training process which is both time-consuming and performance-sensitive. 
In order to fully explore the intrinsic connection between the HR and LR domains, we decide to involve the HR and LR face images equally, expecting to learn a common feature space that is able to cluster LR and HR faces within the same subject as well as maintaining low inter-class proximity despite the difference in resolution. We propose an approach based on \cite{wen2016discriminative}, which employs a novel regularization term to further force the features from the same subject to cluster, which yields better discriminative features. To further stabilize the training process and reduce the overfitting of the model on smaller datasets, we also employ the L2 regularization term. The loss function is:
\begin{equation*}
L=-\sum_{i=1}^{m}\log\frac{e^{W_{yi}^{T}x_{i}+b_{yi}}}{\sum_{j=1}^{n}e^{W_{j}^{T}x_{i}+b_{j}}}+\frac{2 }{\lambda}\sum_{i=1}^{m}\left \| x_{i}-c_{yi}\right \|_{2}^{2}+\eta \left \| W \right \|_{2}^{2}
\end{equation*}
\noindent where $x_{i}$ represents batches of face images from different resolutions, and $c_{yi}$ is the center of each class which updates while training.

\subsubsection{Experiments and Results}

In most of the previous studies, because of the lack of LR images collected from the wild, HR datasets like Multi-PIE \cite{gross2010multi} and FERET \cite{phillips1998feret} are used for this task with downsampling of the HR face images to create  LR counterparts.  Due to the difference between synthetic LR images created by this technique and real surveillance
images, the method has limitations when applied to the real world. Two experiment protocols are defined as follows: 
\begin{itemize}
  \item {\em Closed-set face identification} is typically treated as a classification problem, and the label is predicted directly from the classification layer when a deep learning method is employed. However, when the subjects used for evaluation do not show up in the training stage, the classification architectures tend to be inflexible enough. This subject-disjoint training and testing protocol are more practical in a real-life scenario.
  \item {\em Open-set face identification} requires a 1 to N match when an individual is present to the system. However, it also requires that the system correctly reject individuals who are not enrolled in the system database. This is very similar to how most surveillance systems work. An individual who randomly shows up in the scene may or may not be in the face database. In this case, the system must correctly reject the probe if the person is not in the database and correctly identify if the person is in the database.
\end{itemize}

Evaluations are conducted on two surveillance quality datasets (SCface and UCCSface) which we consider both more challenging and closer to real scenarios. For the network architecture, we employ a shallower network with seven weight layers, described as  conv(32)-conv(32)-maxpooling-conv(64)-conv(64)-maxpooling-conv(128)-conv(128)-maxpooling-flatten-dense. The convolutional kernel size is kept as 3. For center loss, we use a dense layer of 512 dimensions, and for large margin softmax loss~\cite{liu2016large}, additive margin softmax \cite{wang2018additive} and L2-constrained softmax loss \cite{ranjan2017l2}, we use a dimension of 256 for the dense layer. For center loss, we set alpha to 0.9. For L2-constrained softmax loss, alpha is set to 5. For large margin softmax loss we set the margin to 1, and for additive margin loss, we set the margin to 0.35.


\noindent $\bullet$ {\bf SCface dataset:} Each of the 130 subjects in the subset of SCface dataset that we use has one mugshot HR face image taken by the high-definition camera and 15 LR face images that are captured by five visible light cameras placed at three different standoff distances (1m, 2.6m, and 4.2m). This yields 2080 faces in total. We conduct two sets of experiments with two different protocols as defined in  \cite{2018:Yang}.
We split the training set and testing set into 80 and 50 subjects separately in a subject-disjoint fashion. For Experiment 1, we use the HR images as gallery images, and those captured at the three standoffs as probe images. In Experiment 2 we chose face images from 1m standoff as gallery images and images from 2.6m and 4.2m standoff as probe images. The other settings are identical to experiment 1. All the HR and LR images are resized to 64x64 for presentation to the network for training and testing.  We conduct the matching using cosine distance of the matching score and the rank-1 rate is reported in \ref{Tab:Experiment1}. We achieve nearly five percent improvement in rank-1 rate in Experiment 1 and 9 percent in Experiment 2 compared to the state-of-the-art under the same protocol. The feature generated by our model is more robust when the gallery and probe images have a large resolution level difference. The performance of other methods drops rapidly when the size of the probe image changes from 1m standoff to 4.2m standoff while employing feature extracted from our model.

\noindent $\bullet$ {\bf UCCSface dataset:} UCCSface is another dataset designed to be close to a real surveillance setting. We follow the experiment setting provided in \cite{wang2016studying} and \cite{sapkota2013large} and study both closed-set and open-set scenarios. For closed-set evaluation, 180 subjects are used and for open set evaluation, we compare our result with the performance reported in \cite{sapkota2013large} with the defined openness at 14.11 percent. When looking at the result of the closed-set evaluation, our method beats the UCCS baseline by nearly 20 percent on rank-1 accuracy and also outperforms the DNN method in  \cite{wang2016studying} by nearly 35 percent on rank-1 rate under the same training and evaluation protocol. For open-set evaluation, we achieve 73.6 percent accuracy when compared to the UCCS face baseline result.

\begin{table}[]
\centering
\caption{Experiment 1:Rank-1 rate on SCface with HD and three standoff distances}
\normalsize
\begin{tabular}{lrrr}
\hline
\hline
Method                         & HD-1m       & HD-2.6m     & HD-4.7m       \\
\hline
SCface  \cite{grgic2011scface} & 6.18\%       & 6.18\%       & 1.82\%         \\
CLPM  \cite{2010:Li}           & 3.08\%        & 4.32\%        & 3.46\%          \\
SSR  \cite{yang2010image}      & 18.09\%       & 13.2\%       & 7.04\%          \\
CSCDN  \cite{2015:Wang}        & 18.97\%       & 13.58\%       & 6.99\%          \\
CCA  \cite{2015:WangYang}      & 20.69\%       & 14.85\%       & 9.79\%          \\
DCA  \cite{haghighat2017low}   & 25.53\%       & 18.44\%       & 12.19\%         \\
C-RSDA  \cite{2017:Chu}        & 18.46\%       & 18.08\%       & 15.77\%         \\
Centerloss \cite{wen2016discriminative}(ours)              & {\bf 31.71\%} & {\bf 20.80\%} & {\bf 20.40\%}   \\
LMsoftmax \cite{liu2016large}(ours)               & { 18.00\%} & { 16.00\%} & { 14.00\%}   \\
AMsoftmax \cite{wang2018additive}(ours)                & { 18.4\%} & {20.8\%} & { 14.80\%}   \\
L2softmax \cite{ranjan2017l2}(ours)                & { 16.8\%} & { 18.8\%} & { 9.2\%}   \\
\hline
\hline
\end{tabular}
\label{Tab:Experiment1}
\end{table}

\begin{table}[]
\centering
\caption{Experiment 2:Rank-1 rate on SCface with 1.0m and 2.6m standoff distances}
\normalsize
\begin{tabular}{lc}
\hline
\hline
Method                                  & 1.0m-2.6m       \\
\hline
CLPM  \cite{2010:Li}                    & 29.12\%           \\
SDA   \cite{zhou2011low}                & 40.08\%           \\
CMFA  \cite{siena2012coupled}           & 39.56\%           \\
Coupled mapping method  \cite{2015:Shi} & 43.24\%           \\
LMCM     \cite{2016:Zhang}              & 60.40\%           \\
Centerloss \cite{wen2016discriminative}(ours)                       & \textbf {69.60\%} \\
LMSoftmax \cite{liu2016large}(ours)                         & {40.4\%} \\
AMSoftmax \cite{wang2018additive}(ours)                         & {46.8\%} \\
L2softmax \cite{ranjan2017l2}(ours)                         & {42.8\%} \\
\hline
\hline
\end{tabular}
\label{Tab:Experiment2}
\end{table}

\begin{table}[ht]
\centering
\caption{Rank-1 rate on UCCSface dataset}
\normalsize

\begin{tabular}{llc}
\hline
\hline
Resolutions             & Method             & Rank-1 rate     \\
\hline
Original vs Original:   & UCCSface           & 78.00\%           \\                 
                        & Centerloss \cite{wen2016discriminative}(ours)  & \textbf{95.10\%}  \\ 
                        & LMSoftmax \cite{liu2016large}(ours)    & {65.8\%}  \\
                        & AMSoftmax \cite{wang2018additive}(ours)    & {60.6\%}  \\
                        & L2Softmax \cite{ranjan2017l2}(ours)    & {86.50\%}  \\
\hline
80$\times$80 vs. 16$\times$16: & DNN                & 59.03\%           \\
(HR vs. LR)                        &  Centerloss \cite{wen2016discriminative}(ours) & \textbf{93.40\%}  \\
                        & LMSoftmax \cite{liu2016large}(ours)    & {64.9\%}  \\
                        & AMsoftmax \cite{wang2018additive}(ours)    & {58.6\%}  \\
                        & L2Softmax \cite{ranjan2017l2}(ours)    & {85\%}  \\
\hline
\hline
\end{tabular}
\label{Tab:UCCSface}
\end{table}

\subsection{Low-resolution face re-identification}
\label{sec:lr_face_reid}

In this section, we explore LR face re-identification and evaluate it on several datasets captured in an unconstrained environment. 
We employ the VBOLO dataset for an in-depth study and the SCface, UCCSface, and MegaFace challenge 2 LR subset for other topical explorations.


\subsubsection{Experiments and Results}


\paragraph{Actor-Disjoint Experiment with selected deep networks}
\label{subsec:facereid}

We explore the literature and are inspired by recent state-of-the-art patch matching approaches that may exhibit robustness to small misalignments in automatic or manual annotations and robustness to different effective resolutions. We also expect the network to handle matching of faces captured by the same or different cameras with different subject standoffs. We exploit four state-of-the-art face matching approaches with basic DNN architectures, and boost them with fully convolutional structures to reduce overfitting on our dataset. Further, to let the network better accommodate resolution changes, we employ a spatial pyramid pooling (SPP) layer, hoping to learn discriminative features and the mapping between a different size of LR faces captured in the surveillance cameras.

\paragraph{Matching protocols}
\label{matching_protocol}
We make two different matching protocols for this dataset. The first protocol is designed to match people in the same camera between different appearances (usually called ``single camera ReID''). This experiment aims to test face identification performance on people from different appearances in the video while clothes are different in the appearances. The second protocol matches images of people acquired from different camera locations. This protocol aims to evaluate the comprehensive performance of the ReID model which includes single-camera and multi-camera person ReID at the same time. In order to increase the matching complexity, we add distractors to the protocol to obtain more non-match pairs.

We train and test several state-of-the-art deep learning patch-matching architectures on our dataset following the two matching protocols. The training and testing sets are disjoint by actor ID in order to mimic the reality that the targets will be unlikely to appear in the training set. Six of the actors are used for training and three for testing. 
Each set of experiments is conducted five times using the random pair sampling procedure below, and the results are averaged.

\paragraph{Training pairing strategy}
  \label{train_pairing_protocol}
Creating pairs and sampling the generated pairs in training for matching is a key step of preprocessing. Since we have on average around 200 faces for each person in each appearance, the numbers of positive and negative pairs are highly unbalanced.  In addition to pairs created using faces in different appearances, we decide to add face pairs from the same appearance in the training data in order to increase the number of training pairs. 
We denote the training set as $T$, and a particular face in $T$ as $t_{ijf}$, where $i \in 1 \ldots n$ is the ID of the actor, $j \in 1 \ldots m$ is the appearance number, and $f$ denotes the frame number. We first randomly shuffle all the faces in order to break the temporal continuity of the frames to avoid getting positive face pairs more often from frames close in time to each other. For each of the faces $T_{ijf}$ with fixed $i$ and $j$, we create a positive pair by randomly selecting $T_{ij'f'}$ with $j' \ne j$ and $f' \ne f$. To create an equal number of negative pairs, we randomly select the faces from $T_{i'j’f'}$ with $i' \ne i$ and $j'$ and $f'$ randomly chosen, and pair the selected face with the previous face $T_{ijf}$. By exploiting the pairing approaches mentioned above, we are able to gain balanced face pairs from each identity in each appearance.

\begin{figure}

    \centering

    \subfloat[]{{\includegraphics[width=8cm]{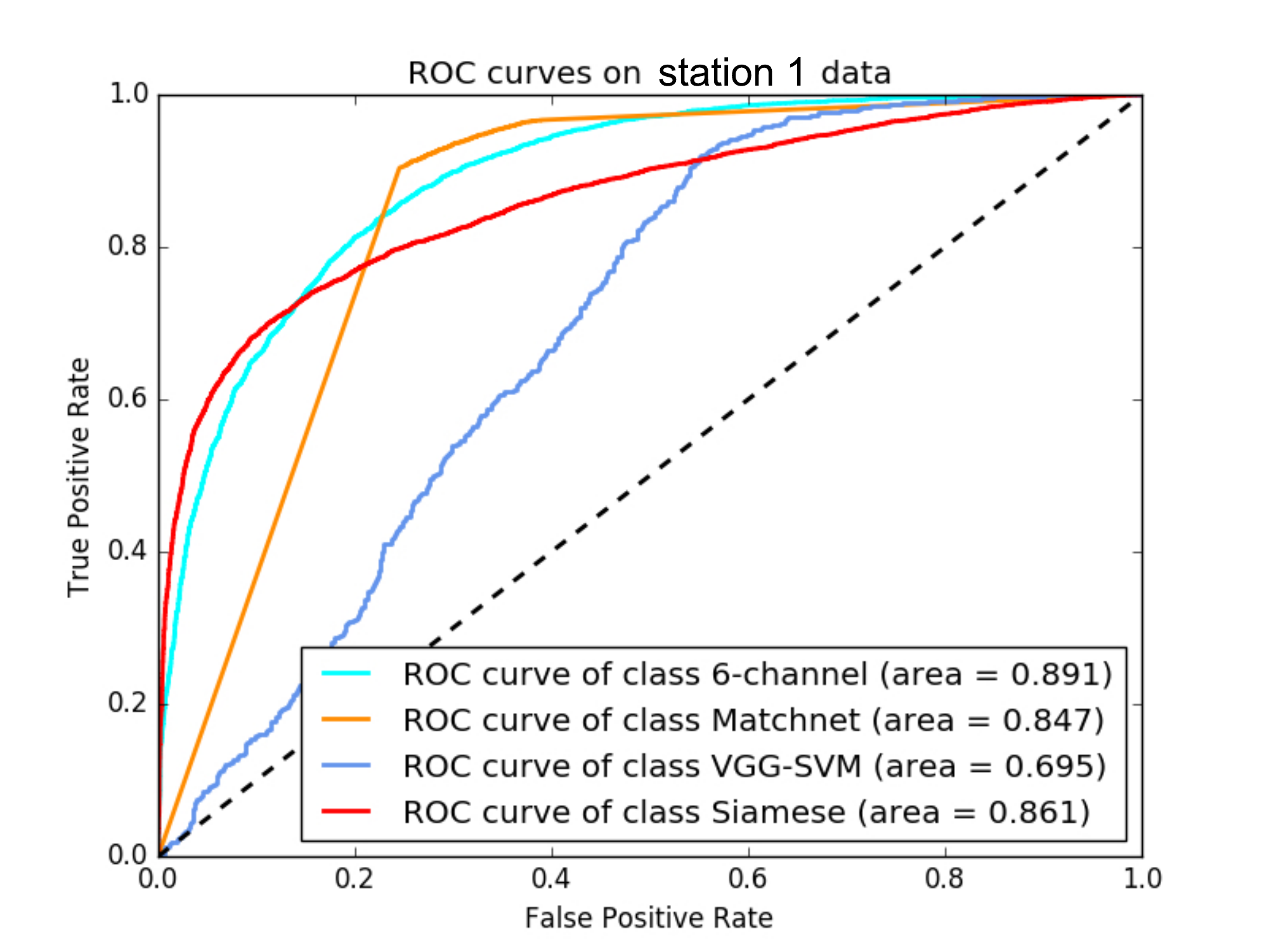} }} \\
    \subfloat[]{{\includegraphics[width=8cm]{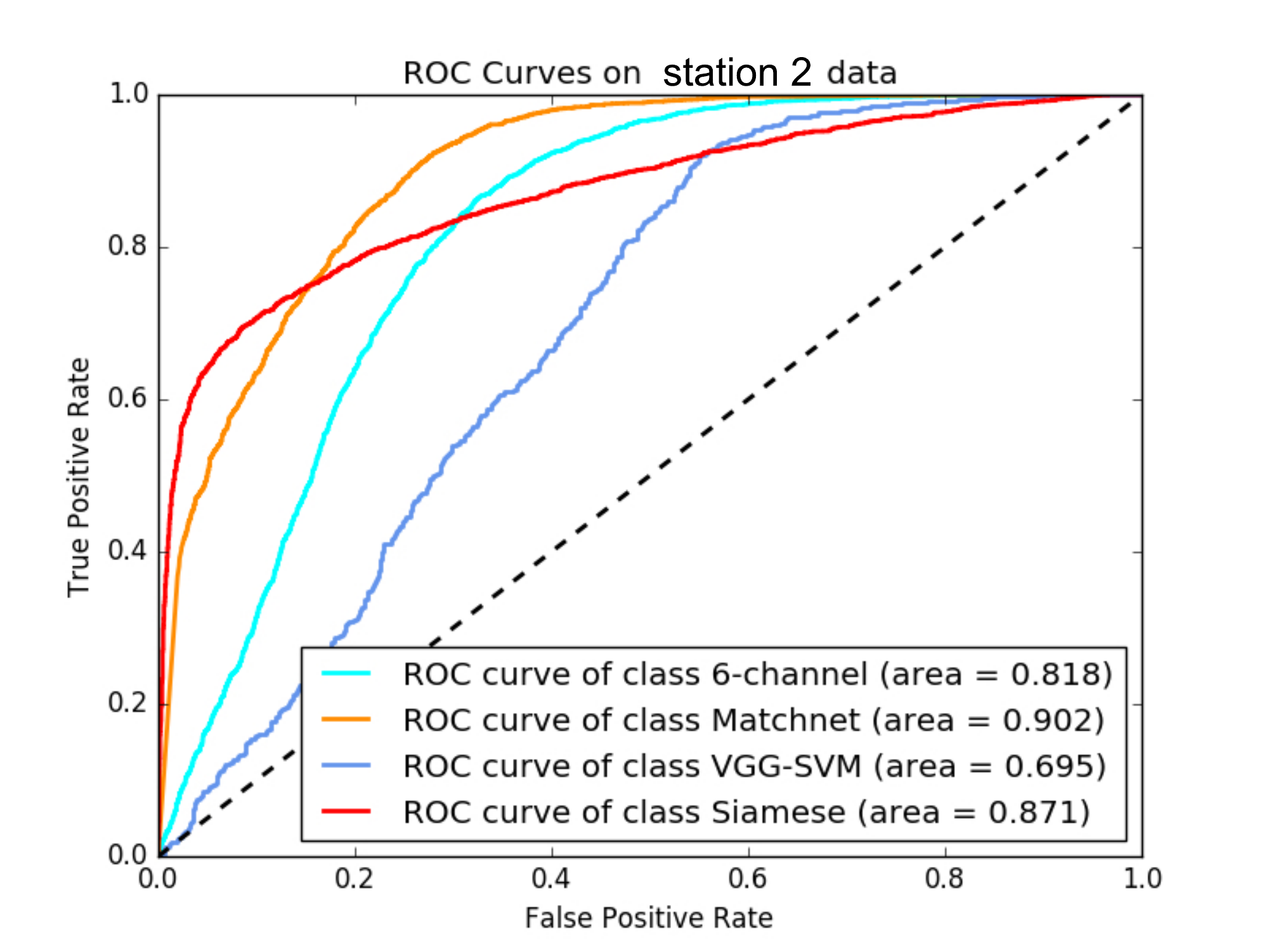} }}\\
    \subfloat[]{{\includegraphics[width=8cm]{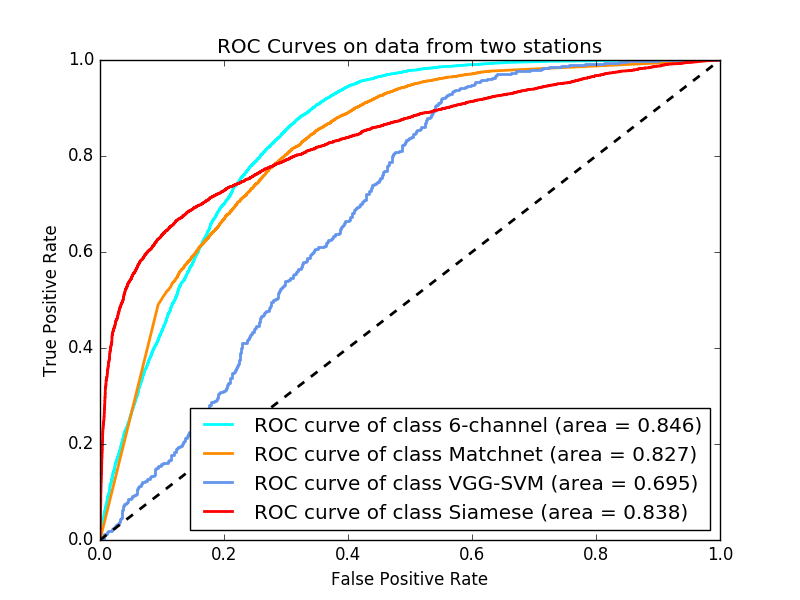} }}
    \caption{Face matching with three matching basic net (first line in Table \ref{Tab:fcnn}) for three data subsets. (a): Station 1 (b): Station 2 (c) Station 1 and Station 2.}
    \label{Fig:BasicMatching}

\end{figure}

\paragraph{VGG-Face \& SVM}
We employ the pre-trained VGG-face descriptor model and use it for facial feature extraction. We pair the faces from our dataset first and extract features using the VGG face descriptor \cite{parkhi2015deep}. The face images are resized from their original size to 224$\times$224 for input to the network.
 The Euclidean distance between feature vectors from a pair of faces is assigned with positive and negative binary labels to represent if the pair of faces is from the same person or not. The distances themselves are used to train a linear SVM model for binary prediction. 
We conduct the experiment with the two matching protocols mentioned above. We chose the hyper-parameter value and obtain the best CV rate for the linear SVM, which yielded
a testing AUC of 0.695 as shown in Fig. \ref{Fig:BasicMatching}. This result indicates the network successfully identified face features from poor quality faces extracted from surveillance videos. However, since the VGG face descriptor is trained on various HR faces that are sufficiently aligned with facial landmarks, when used for surveillance quality faces that are both LR and hard to align, this baseline result is not outstanding. Also, LR face images need to be upscaled by a large factor before being fed into the deep pre-trained network, which may introduce artifacts and also increase computational complexity.

\paragraph{Siamese Network} 
Siamese classification structures had their first application in face verification in \cite{chopra2005learning}. The Siamese architecture does not require categorical information in training. It tries to learn a feature representation with two identical towers of network layers with shared weights. It has a series of convolutional, activation and max-pooling layers in each tower. Two feature representations coming from each tower are concatenated and fed into fully connected layers which are connected with contrastive loss. Since the two towers are identical in structure and weights, this kind of network aims to map two inputs into an identical target space by training in an end-to-end fashion. We propose a tiny base network with a simple architecture, motivated by \cite{wang2016studying}, which concludes that a deeper network architecture and a large number of filter channels may degrade recognition performance. Our basic network is shown in Table \ref{Tab:fcnn}; it has three convolutional layers followed by max-pooling and a fully connected layer. We use a moderate filter size and channel number in the tiny network. An input size of 32$\times$32 is chosen. We train the network from scratch with a batch size of 8 and the SGD optimizer. The Siamese net converged within 5 epochs with an AUC of 0.861 on Station 1 data and 0.871 on Station 2 data with single camera matching and 0.838 on the data from two stations with both single and cross camera matching. 

\begin{figure}

    \centering
    
    \includegraphics[width=7.5cm]{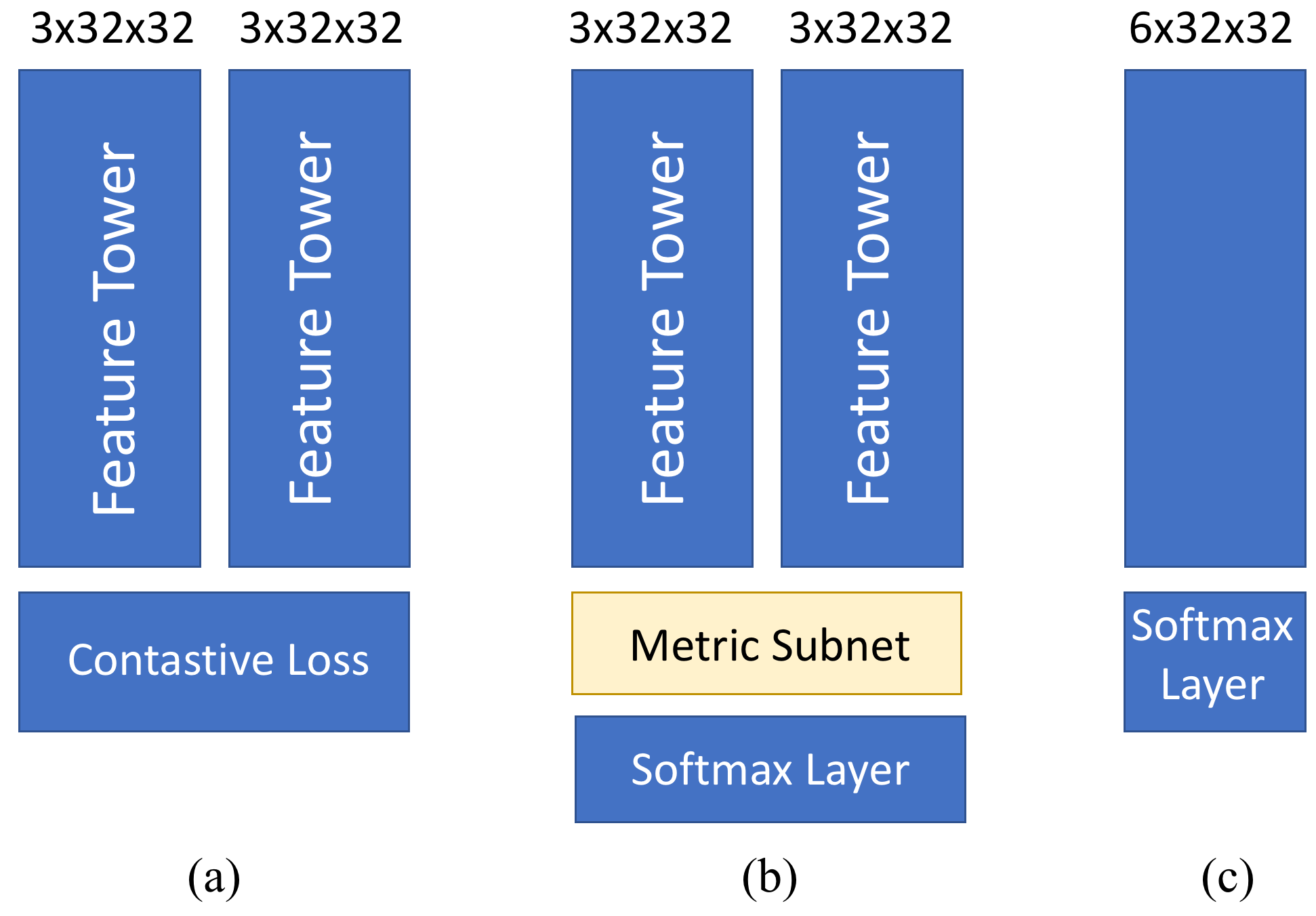} 
    \caption{Overview of the three deep architecutures: (a)Siamese net   (b) Matchnet (c) 6-channel net }
    \label{Fig:DeepArchitectures}

\end{figure}

\paragraph{MatchNet} 
 MatchNet \cite{han2015matchnet} is another state-of-the-art patch matching approach that employs a two-tower structure with shared weights similar to the Siamese net. However, instead of feeding two concatenated feature vectors produced by the two towers directly into the decision layer with carefully designed loss functions, it uses a series of fully connected layers as a subnet to learn the feature comparison for binary classification using cross-entropy. Compared to the Siamese net, MatchNet has more flexibility in the metric subnet shown in Fig. \ref{Fig:DeepArchitectures}, which takes the paired features and maps them to a unified space that minimizes their distance. 
 However, it converges more slowly since the fully connected layer has many more parameters and higher complexity.  A softmax layer and cross-entropy loss are employed during training. We obtain the best result on our basic net
using the SGD optimizer. As shown in Fig. \ref{Fig:BasicMatching}, it obtains AUC of 0.847 on data from Station 2, 0.902 on data from Station 1 for single camera matching, 0.827 on data from both stations for single and cross-camera matching. 

\paragraph{Six-Channel Net} 
Inspired by the two-channel model proposed in \cite{zagoruyko2015learning}, we improve it by incorporating three color channels. This approach abandons the two-tower feature by directly embedding the two face images into six channels, fed into the first layer of the network, with hinge loss and a one-bit binary output which is shown in Fig. \ref{Fig:DeepArchitectures}. 
Compared to the previous two architectures, it has greater flexibility--it has twice as many parameters as the two-tower structures and is able to learn feature maps using six image channels instead of three jointly. However, it converges slowest among the three and L2 regularization is needed for better performance. The best AUC values we achieve using the six-channel net on Station 1 and Station 2 single camera matching are 0.891 and 0.818. It achieves AUC of 0.846, which outperform the Siamese net and MatchNet by 2 percent and 1 percent, respectively.



\begin{table}

\centering
\caption{Model selection for a convolution kernel size of 3\rx 3. Testing accuracy is shown with different network layouts}
\small
\begin{tabular}{llc}
\hline
\hline
Parameters &Method&Accuracy\\
\hline
c(32)-m-c(32)-m-c(64)-fc(64)           & 6-channel & 77.1\% \\ 
&MatchNet&77.3\% \\
&Siamese&78.3\% \\
\hline
c(8)-m-c(16)-m-c(32)-c(32)-m           & 6-channel & 76.7\% \\ 
&MatchNet&78.1\% \\
&Siamese&77.8\% \\
\hline
c(8)-m-c(16)-c(16)m-c(32)-c(32)-m      & 6-channel & 77.5\% \\ 
&MatchNet&79.3\% \\
&Siamese&79.8\% \\
\hline
c(8)-c(8)-m-c(16)-c(16)m-c(32)-c(32)-m & 6-channel & 76.8\% \\ 
&MatchNet&75.9\% \\
&Siamese&78.5\% \\
\hline
\hline
\end{tabular}
\label{Tab:fcnn}
\end{table}

    

\begin{figure}
    \centering
    \includegraphics[width=4cm]{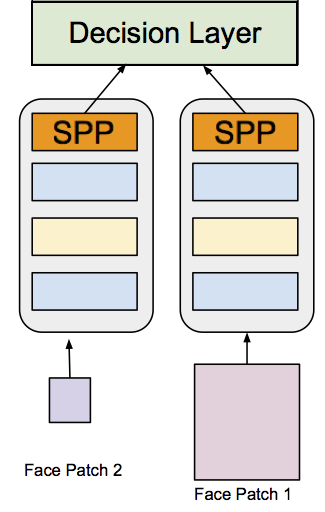}
    \caption{Architecture of Siamese network with SPP layer.}
    \label{Fig:SiameseArchitecture}
\end{figure}

\paragraph{Fully convolutional structure and SPP pooling}

\begin{figure*}[]

\begin{center}

    \subfloat[]{{\includegraphics[width=0.4\textwidth]{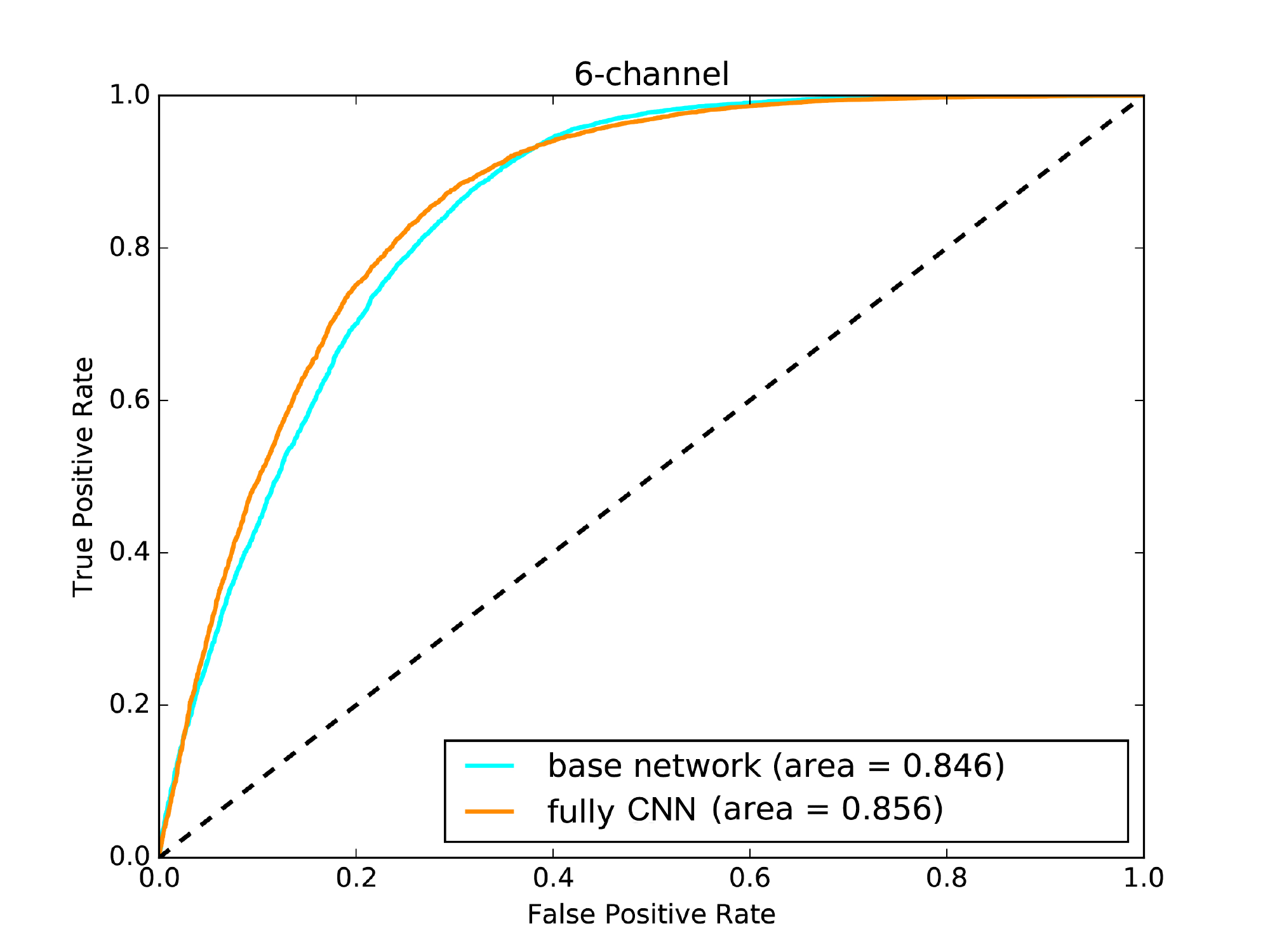} }}
    \subfloat[]{{\includegraphics[width=0.4\textwidth]{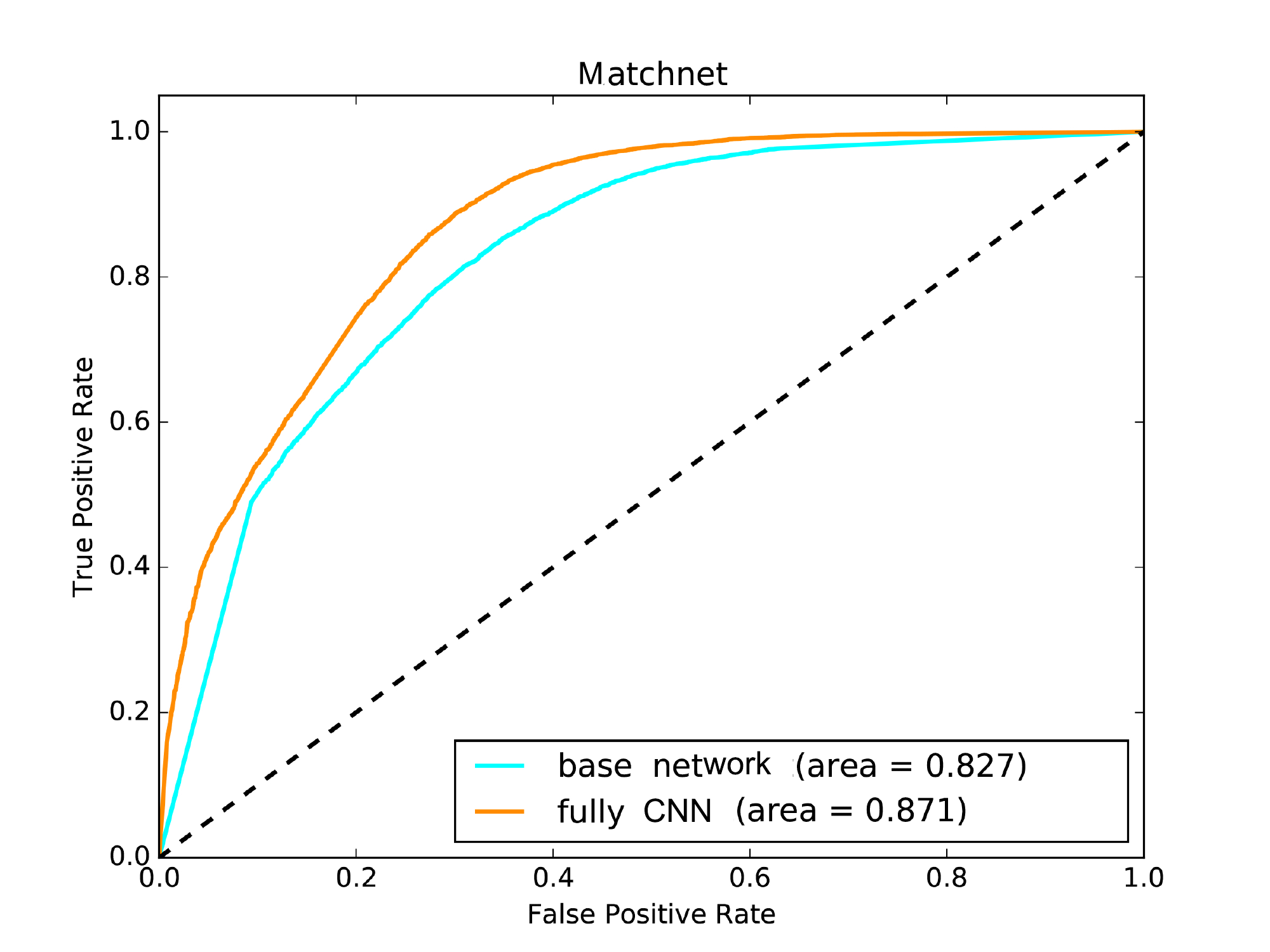} }}
    \\
    \subfloat[]{{\includegraphics[width=0.4\textwidth]{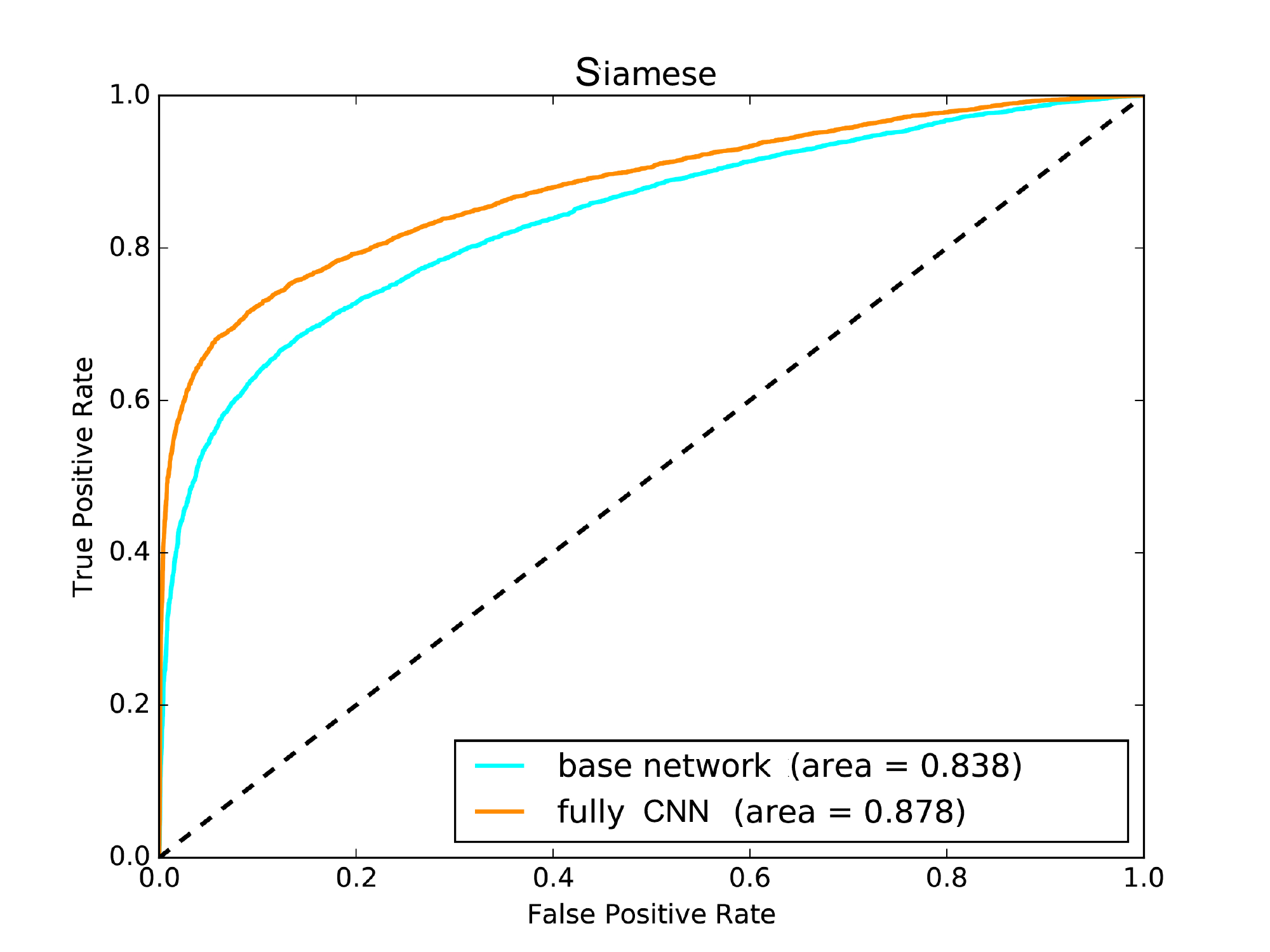} }}
    \subfloat[]{{\includegraphics[width=0.4\textwidth]{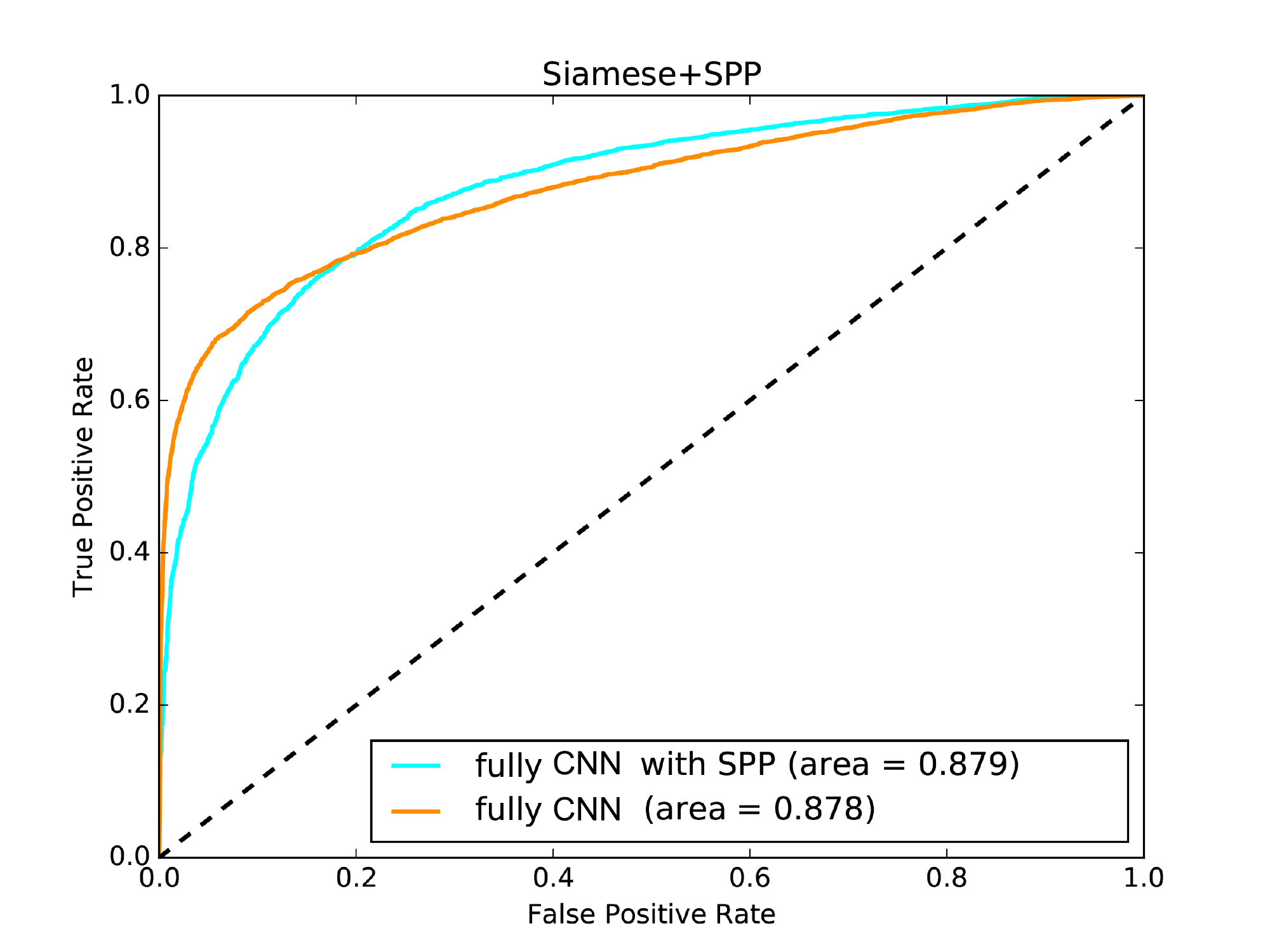} }}
    
    \caption{Face matching with fully convolutional networks on faces from two stations with multi-camera and  multi-appearances. (a): 6-channel.  (b): Matchnet.   (c): Siamese net. (d): Siamese new with SPP layer.}
    \label{Fig:FCNN_Evaluation}
\end{center}
\end{figure*}

In this section, we try to improve the three architectures mentioned above (the Siamese net, MatchNet, and the 6-channel net with a fully convolutional structure and a Spatial Pyramid Pooling (SPP) layer  \cite{kaiming14ECCV}). A fully convolutional CNN (FCN) is one where all the learnable layers are convolutional. A convolutional layer has fewer parameters than a fully connected layer, which would potentially reduce overfitting on a small dataset but keep more spatial information in the features. We replace the fully connected layer with a convolutional layer in the previous three models, test several hyperparameter settings to adjust layer numbers and filter numbers, and chose the best settings based on the testing accuracy observed. Performance comparison is summarized in Table  \ref{Tab:fcnn}. Fig. \ref{Fig:FCNN_Evaluation} demonstrates that the FCN architecture effectively improves the performance (as measured by the AUC) over the three basic network architectures (Siamese, Matchnet, 6-channel net) 
by 1 percent, 5 percent and 4 percent roughly each.
Further, we are inspired by the study of Zagoruyko et al. \cite{zagoruyko2015learning} who used an SPP layer for patch matching and claimed a remarkable improvement in performance. Compared with  \cite{zagoruyko2015learning} who only tested their architecture with SPP using size-identical image pairs, we decide to take advantage of the fully connected architecture of our network, feeding various size of face images into the network. By replacing the last max-pooling layer with an SPP layer before the decision layer, we can further test the assumption from \cite{zagoruyko2015learning} on the VBOLO dataset with our modified architecture. We simplify the problem by setting up three convolutional Siamese networks. Each is responsible for a matching at a specific resolution level. In this case, we have a) low to low, b) high to high, and c) low to high-resolution matching with three separate but identical networks. These three networks try to learn different metrics and features with face pairs close to their original sizes. We resize faces that are smaller than 32\rx 32 to 16\rx 16 and denote them as LR faces. Those faces bigger than 32\rx 32 are resized to 64\rx 64 and denoted as HR faces. We train and test the three subnets with the SPP layer on the top shown in Fig. \ref{Fig:SiameseArchitecture}. 4\rx 4 SPP pooling is applied at the end of each tower. We got a slight (0.1 percent) AUC improvement using the SPP layer together with the fully convolutional architecture. Compared with previous work \cite{li2016toward}, 
we achieve a significant improvement in performance on the data from Station 2 by exploiting the unified deep feature and metric learning instead of optimizing feature and metric separately. 

\begin{figure*}[]

    \centering
    \includegraphics[width=17cm]{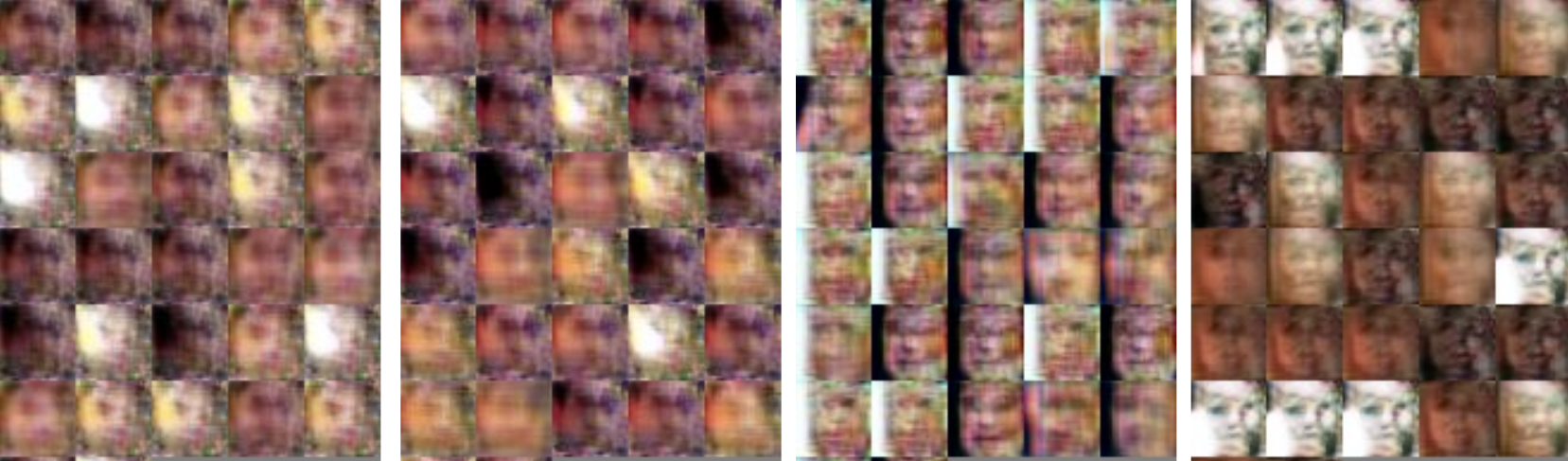}

    \caption{Generated LR faces from DCGAN generator }
    \label{DCGAN}
\end{figure*}

\paragraph{Improved pretraining aproach on larger datasets}
Although we achieve several sets of promising results on the VBOLO dataset, more comprehensive experiments need to be conducted on larger scale public datasets. In addition, we face the challenge that most of the deep architectures struggle to exploit LR images well due to over-fitting and limited intrinsic dimensionality of the input. We execute a project to improve the training process by evaluating on some larger general LR unconstrained face datasets or surveillance quality face datasets. To achieve this goal, we employ the DCGAN introduced in \cite{radford2015unsupervised} in order to obtain a pre-trained discriminator as an initialization for the feature towers.
Using this method has two advantages:
\begin{itemize}
    \item 
     By optimizing the DCGAN on the LR training set, we can understand how the network perceives the LR images and adjust parameters such as activation function and the number of layers by looking at the intermediate output of the generator as shown in Fig. \ref{DCGAN}. Visualizations from filters and feature maps are not yet intuitive enough to inform a training strategy.
    \item
    The pretrained discriminator could provide an initial weight on general LR faces which can be transferred to other LR face datasets via fine-tuning which can stabilize and accelerate the training process.
\end{itemize}
The GAN discriminator is trained on the MegaFace challenge 2 LR subset and fine-tuned using the target datasets (VBOLO, SCface and UCCSface). We compare the model trained from scratch and a fully convolutional MatchNet model pre-trained using DCGAN.

\mbox{}

We perform the training and testing splitting as follows: For UCCSface, we choose 90 identities for training and 90 for testing. For MegaFace challenge 2 LR subset, we use 2999 identities for training and 6699 identities for testing. For SCface, we select 65 identities for training and 65 for testing. We follow the VBOLO pairing and matching protocol for training and testing. The validation set is a randomly selected 20 percent of the training set. All the experiments are run five times 
and the error rates are averaged. When starting with a pretrained weight using a larger number of LR faces by applying the DCGAN discriminator, we observe that the validation rate is more stable and tended to be higher compared with training the network from scratch. Better performance is achieved on the same datasets. For VBOLO, UCCSface, MegaFace challenge 2 LR subset and SCface, we achieve 9.2, 21.8, 10.6, 11.3 percent decrease in error rate compared with training the model from scratch.

We also identify some architectural changes needed for a DCGAN model to successfully converge on LR face images compared with on HR images. For higher resolution modeling, Radford et al. \cite{radford2015unsupervised} suggested that the following steps would result in stable training:
\begin{itemize}
    \item Replace any pooling layers with strided convolutions (discriminator) and fractional-strided convolutions (generator).
    \item Replace Tanh activation function with ReLU or Leakyrelu functions.
    \item Add batch-normalization in both generator and discriminator.
\end{itemize}
However, we only achieve stable adversarial convergence using the Tanh activation function in both the generator and the discriminator except for the last layer of the discriminator, in which we employ a sigmoid nonlinearity. Batch-normalization does not function usefully for our DCGAN to converge on the MegaFace LR subset and is not applied in our model.
 
\begin{table}[t!]
\centering
\footnotesize
\caption{Average error rate on Matchnet models}
\begin{tabular}{lcccc}
\hline
\hline
Datasets           & VBOLO     & UCCSface  & MegaFace 2 sub  & SCface    \\ 
\hline
DCGAN-pretrained   & 18.8/17.6 & 14.7/11.8 & 20.1/19.8 & 24.3/24.1 \\ 
Train from scratch & 20.7/19.5 & 18.8/18.6 & 22.5/21.8 & 27.4/28.5 \\ 
\hline
\hline
\end{tabular}
\label{Tab:MatchnetError}
\end{table}

\section{Conclusions and Commentary}
\label{Sec:Conclusions}
%
In this paper, we provide several novel contributions.
First, we illustrate the performance gap between LR unconstrained face and LR constrained face recognition when using a state-of-the-art super-resolution algorithm. Secondly, two important application scenarios based on LR face recognition are defined:  unconstrained LR face identification in the wild and LR face re-identification. For general LR face identification, we exploit a novel approach to handle the multi-dimensional mismatching due to the quality difference of the face images in probe and gallery. We also design different deep networks solving the person re-identification problem to demonstrate better performance compared to our previous work \cite{li2016toward,li2017learning}. We present a novel strategy using DCGAN pre-training to obtain both the learning visualization of the network and improve result on larger scale datasets. We demonstrate the result from the extensive experiments on selected datasets and discover that dimensional mismatching is the most challenging point, especially in low-to-high resolution face identification task. The result demonstrates that the approaches we proposed target different tasks, work efficiently, and yield impressive performance.


\bibliographystyle{ieee}
\bibliography{tifs.bib}

\end{document}